\documentclass[conference]{IEEEtran}
\usepackage{cite}
\usepackage{amsmath,amssymb,amsfonts}
\usepackage{algorithmic}
\usepackage{graphicx}
\usepackage{textcomp}
\usepackage{xcolor}
\usepackage{subcaption}

\usepackage{graphicx}   % already in most templates
\usepackage{booktabs}   % for \toprule, \midrule, \bottomrule
\usepackage{caption}    % for \captionof

\IEEEoverridecommandlockouts

\usepackage{tikz}
\usepackage{xcolor}
\usepackage{float}

\newcommand{\hw}[1]{\ensuremath{\mathtt{#1}}}

\usepackage{booktabs} % for better-looking tables
\usepackage{array} % for customizing column widths
\usepackage{xspace}

\newcommand*\ding[1]{\tikz[baseline=(char.base)]{
            \node[shape=circle,fill,inner sep=1pt] (char) {\footnotesize \textcolor{white}{#1}};}}

\usepackage{algorithm}
\usepackage{algorithmic}
\usepackage[]{hyperref}
\def\BibTeX{{\rm B\kern-.05em{\sc i\kern-.025em b}\kern-.08em
    T\kern-.1667em\lower.7ex\hbox{E}\kern-.125emX}}
\begin{document}
\title{\huge GROOT: \underline{G}raph Edge \underline{R}e-growth and Partiti\underline{o}ning for the Verification of Large Designs in L\underline{o}gic Syn\underline{t}hesis\\}
\author{%
Kiran~Thorat$^{1}$,
Hongwu~Peng$^{1}$,
Yuebo~Luo$^{3}$,
Xi~Xie$^{1}$,
Shaoyi~Huang$^{1}$,
Amit~Hasan$^{1}$,\\
Jiahui~Zhao$^{1}$,
Yingjie~Li$^{2}$,
Zhijie~Shi$^{1}$,
Cunxi~Yu$^{2}$,
Caiwen~Ding$^{3}$\\[2pt]
$^{1}$University of Connecticut\quad
$^{2}$University of Maryland\quad
$^{3}$University of Minnesota\\[2pt]
$^{1}$\{kiran\_gautam.thorat, hongwu.peng, xi.xie, shaoyi.huang, amit.hasan, jiahui.zhao, zhijie.shi\}@uconn.edu\\
$^{2}$\{yingjiel, cunxiyu\}@umd.edu\quad
$^{3}$\{luo00466, dingc\}@umn.edu
\thanks{This work was supported in part by the National Science Foundation under IUCRC-1916762, and CHEST (Center for Hardware Embedded System Security and Trust) industry funding.}%
}
\maketitle
\begin{abstract}
% \begin{abstract}
% Verification is an integral part of logic synthesis in chip design, that ensures the resulting hardware meets the intended functionality. 
Traditional verification methods in chip design are highly time-consuming and computationally demanding, especially for large scale circuits. Graph neural networks (GNNs) have gained popularity as a potential solution to improve verification efficiency. However, there lacks a joint framework that considers all chip design domain knowledge, graph theory, and GPU kernel designs.
% an effective EDA-based graph learning solution requires a fusion of EDA domain expertise and profound knowledge of graph machine-learni, 
% especially when dealing with large circuits. 
To address this challenge, we introduce GROOT, an algorithm and system co-design framework that contains chip design domain knowledge and redesigned GPU kernels, to improve verification efficiency. More specifically, we create node features utilizing
the circuit node types and the polarity of the connections between the input edges to nodes in And-Inverter Graphs (AIGs). We utilize a graph partitioning algorithm to divide the large graphs into smaller sub-graphs for fast GPU processing and develop a graph edge re-growth algorithm to recover verification accuracy.
% to  based on the observation that approximately only 10\% boundary edges (nodes) between cluster, to divide the large graphs into smaller sub-graphs for fast GPU processing.
We carefully profile the EDA graph workloads and observe
the uniqueness of their polarized distribution of high degree (HD) nodes and low degree (LD) nodes. We redesign two GPU kernels (HD-kernel and LD-kernel), to fit the EDA graph learning workload on a single GPU.
% combines graph partitioning and edge re-growth methods to significantly reduce memory requirements for large circuit design graphs. 
% We evaluate the performance of GROOT on large circuit designs, e.g., Carry Save
% Adder (CSA) multipliers, the 7nm technology mapped CSA multipliers, FPGA mapped CSA multipliers, and Booth Multipliers.
We compare the results with state-of-the-art (SOTA) methods: GAMORA, a GNN-based approach, and the traditional ABC framework.
% We compare the results with state-of-the-art (SOTAs), GAMORA (a GNN-based) and the traditional ABC framework. 
% \cd{add kernel design.} 
Results show that GROOT~achieves a significant reduction in memory footprint (59.38 \%), with high accuracy (99.96\%) for a very large CSA multiplier, i.e. 1,024 bits with a batch size of 16, which consists of 134,103,040 nodes and 268,140,544 edges. We compare GROOT~with GPU-based GPU Kernel designs SOTAs such as 
% will be illustrated in Fig. \ref{fig:eda_kernel_speedup}, comparing them to 
cuSPARSE, MergePath-SpMM,
% the academical frameworks such as 
and GNNAdvisor.  We achieve up to 1.104$\times$, 5.796$\times$, and 1.469$\times$ improvement in runtime, respectively.
 
% \end{abstract}
\end{abstract}

% \begin{IEEEkeywords}
% \color{red}
% component, formatting, style, styling, insert
% \end{IEEEkeywords}

\section{Introduction}
%\section{Introduction}
% In the process of chip design, 
Logic synthesis plays a vital role in chip design by converting high-level circuit descriptions into optimized gate-level implementations and helps to bridge the gap between high-level synthesis and physical design \cite {JIANG2009299}. Verification is a critical step in logic synthesis that ensures internal functionality, prevents costly errors, and reduces the time-to-market by identifying and fixing issues early in the design cycle~\cite{10.1145/3543853}.
% [REF])}
% It is performed at different synthesis stages and ensures that the final hardware meets the intended functionality . 
However, traditional verification methods are computationally demanding and increasingly time-consuming, especially for complex designs \cite{Kaufmann+2022+285+291, 8105885}.
For example, as measured in \cite{7459464}, the verification process takes more than 100 hours for the booth multiplier
% , and more than  1 hour for the 128-bit multiplier utilizing ripple carry adder, 
% the verification times for various 128-bit multiplier implementations 
using the OneSpin commercial equivalence checker tool.
Furthermore, using the open-source verification tool ABC \cite{Mishchenko2007}, a 2048-bit multiplier requires \(8.6 \times 10^5\) seconds (more than nine days) \cite{wu2023gamora}.
Graph neural networks (GNNs) have gained popularity as a potential solution to improving verification efficiency, e.g., GAMORA~\cite{wu2023gamora}, since graph is one the natural ways to represent many fundamental objects in circuits, such as Register Transfer Level (RTL) descriptions, netlists, layout, and Boolean functions. In GNN-based methods, GNN is leveraged to classify the graph nodes which significantly reduces the verification time. For example, the 2048-bit multiplier verification time reduces from more than nine days to 0.919 seconds when GNN is used \cite{wu2023gamora}.

\begin{figure}[t]
\centering
\begin{minipage}{0.48\linewidth}
    \centering
    \includegraphics[clip,width=\linewidth]{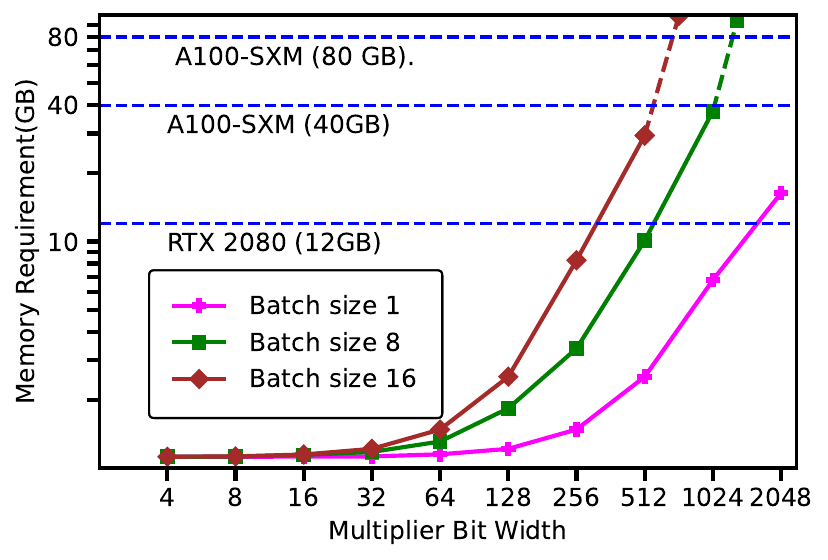}
    \text{(a)}
\end{minipage}%
\hspace{0.01\linewidth}% Adjust space between figure and table
\begin{minipage}{0.48\linewidth}
    \centering
    \begin{table}[H]  % Encapsulate the table in a table environment
        \centering
        \resizebox{\linewidth}{!}{
        \begin{tabular}{lccc}
            \toprule
            \textbf{Methods} &
            \begin{tabular}[c]{@{}c@{}}\textbf{Large}\\\textbf{multipliers}\end{tabular} &
            \begin{tabular}[c]{@{}c@{}}\textbf{Run}\\\textbf{time}\end{tabular} &
            \begin{tabular}[c]{@{}c@{}}\textbf{Accu}\\\textbf{racy}\end{tabular} \\
            \midrule
            \textbf{ABC~\cite{Mishchenko2007}} & \checkmark & $\times$ & \checkmark \\
            \midrule
            \textbf{GAMORA~\cite{wu2023gamora}} & $\times$ & \checkmark & \checkmark \\
            \midrule
            \textbf{GROOT~(ours)} & \checkmark & \checkmark & \checkmark \\
            \bottomrule
        \end{tabular}}
        \label{table-comparison}
    \end{table}
    \text{(b)}
\end{minipage}
\caption{(a) Extremely high GPU memory requirements on large circuit graphs in EDA. CSA multiplier with different bits and batch sizes, (b) Comparison of verification methods.}
\label{fig:memory_edited}
\end{figure}

Despite their promising results, there are research gaps.
% exist in GNN-based EDA research. 
\textbf{First}, an effective graph machine-learning solution for logic synthesis requires a fusion of electronic design automation (EDA) domain expertise and knowledge of graph machine learning. However, existing efforts tend to focus on just one aspect, such as applying GNN algorithms to EDA tasks, and may lack EDA domain expertise. For instance, GAMORA~\cite{wu2023gamora} does not distinguish Primary Inputs and Primary Outputs (PO) when creating graph node features, however, {PI} and {PO} are inherently different and need to be distinguished.
\textbf{Second}, processing a large-scale EDA GNN on a single hardware, which is crucial to efficient AI, has been largely neglected. Figure \ref{fig:memory_edited} (a) shows the memory consumption (on two high-end GPUs NVIDIA A100 40 GB and 80 GB, and one low-end GPU GeForce RTX2080) required for the verification of various bit widths multipliers. We observe that even the NVIDIA A100 could not accommodate
% the multiplier graph for 
the 1,024-bit CSA multiplier graph when batch size equals 16.
% Specifically, 
% when batch processing is used, 
% GeForce RTX2080 is unable to handle multipliers of bit widths greater than 256 bits when the batch size is 16. 
Please note that batch processing is essential to achieve high throughput 
 % it is essential to employ batch processing 
 as GPUs are designed to process parallel data. \textbf{Third}, the state-of-the-art (SOTA) high-performance solutions often use GPU, and simply adopt commercialized  multi-GPU solutions (e.g., GAMORA directly uses Pytorch Geometric~\cite{fey2019fast} on two or more GPUs). However, an important aspect that frequently goes unnoticed is the consideration of GPU primitives. This fundamentally limits making single GPU achievable for EDA GNN and the applicability of broadening accessibility in economically disadvantaged districts.

In this research, we propose GROOT, \underline{G}raph Edge \underline{R}e-growth and Partiti\underline{o}ning for the Verification of Large Designs in L\underline{o}gic Syn\underline{t}hesis. GROOT is a single-GPU-based framework and simultaneously achieves high accuracy, and low memory footprint at run-time.  
The classical open-source EDA tool ABC~\cite{Mishchenko2007}  is not capable of obtaining verification results at run-time, and 
GAMORA~\cite{wu2023gamora} faces the out of  memory issue on large circuit graphs, as summarized in figure \ref{fig:memory_edited} (b).

Our key contributions are: 

\begin{itemize}
    \item At the EDA domain level, we revisit and redesign the node features for the EDA graphs and create datasets. We use circuit node types and the polarity of connections between input edges and nodes in And-Inverter Graphs (AIGs) to form the input embedding. Adding more features to the EDA graphs helps the GNN model learn broader characteristics of designs.
    
    \item At the graph processing level, we use EDA domain knowledge and a graph partitioning algorithm to split large graphs into smaller sub-graphs for GPU processing. Further, we develop a boundary edge re-growth algorithm for accuracy recovery.

    \item We carefully profile the EDA graph workloads and note the polarized distribution of high degree (HD) nodes and low degree (LD) nodes. We redesign two GPU kernels (HD-kernel and LD-kernel) to accelerate the EDA graph learning workload on a single GPU.
\end{itemize}
 Experimental results show that GROOT~achieves a significant reduction in memory footprint (59.38 \%), with high accuracy (99.96\%)  for large circuit graphs, compared to ABC and GAMORA. Note that the accuracy of node classification directly
translates to the verification accuracy. Compared to SOTA GPU-based GPU Kernel designs such as 
% will be illustrated in Fig. \ref{fig:eda_kernel_speedup}, comparing them to 
cuSPARSE \cite{naumov2010cusparse}, MergePath-SpMM~\cite{shan2023mergepath},
% the academical frameworks such as 
and GNNAdvisor~\cite{wang2021gnnadvisor}, GROOT~achieves up to 1.104$\times$, 5.872$\times$, and 1.469$\times$ improvement in runtime, respectively.
% ACM's consolidated article template, introduced in 2017, provides a
% consistent \LaTeX\ style for use across ACM publications, and
% incorporates accessibility and metadata-extraction functionality
% necessary for future Digital Library endeavors. Numerous ACM and
% SIG-specific \LaTeX\ templates have been examined, and their unique
% features incorporated into this single new template.

% If you are new to publishing with ACM, this document is a valuable
% guide to the process of preparing your work for publication. If you
% have published with ACM before, this document provides insight and
% instruction into more recent changes to the article template.

% The ``\verb|acmart|'' document class can be used to prepare articles
% for any ACM publication --- conference or journal, and for any stage
% of publication, from review to final ``camera-ready'' copy, to the
% author's own version, with {\itshape very} few changes to the source.

 \section{Background and Related Work}
\textbf{Verification in EDA}
Verification
% plays an integral role in logic synthesis and/
can be performed at multiple stages 
% (Figure~\ref{fig: chip design})
to ensure that the designed chip meets its intended functionality. 
% However, verification tasks are expensive and complex, primarily due to coverage requirements.
% \begin{figure*}[!h]
% \centering
% \includegraphics[width=.9\textwidth]{Figures/Overview_updated.pdf}
%  %\vspace{1mm}
%     \caption{ Overview of the tasks: (a) Circuit to Transitional Graph Conversion, (b) Graph Input Embedding and Node Label generation based on transitional graphs, (c) Large Graph Partitions to Solve GPU memory issues, (d) Graph Neural Network Architecture, (e) Node classification.} 
%     \label{fig:Overview_flow}
%     % \vspace{-0.1in}
% \end{figure*}

\begin{figure*}[!t]
\centering
\includegraphics[width=\textwidth]{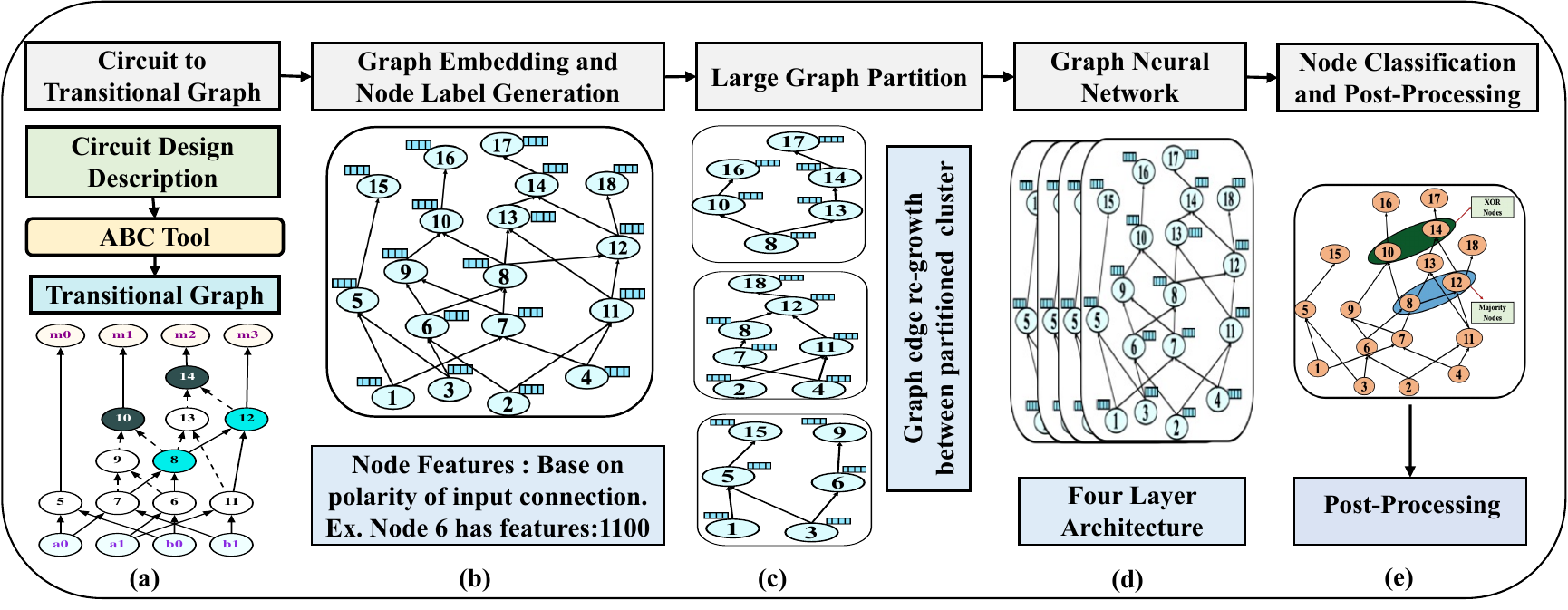}
\caption{Overview of the tasks: (a) Circuit to Transitional Graph Conversion, (b) Graph Input Embedding and Node Label generation based on transitional graphs, (c) Large Graph Partitions to Solve GPU memory issues, (d) Graph Neural Network Architecture, (e) Node classification.}
\label{fig:Overview_flow}
\end{figure*}

Traditional formal verification techniques include Satisfiability (SAT), canonical diagrams, theorem proving \cite{7167236}, and algebraic re-writing. The SAT technique models the verification problem as Boolean satisfiability \cite{7442835,8894250}. Canonical diagrams propose different graph-based representations, such as binary decision diagrams (BDDs) \cite{1676819}, 
Taylor expansion diagrams (TEDs) \cite{1668046}, and binary moment diagrams (BMDs) \cite{1586761}.
%\cd{where are theorem proving related references?}
The algebraic approaches, based on modeling circuit specifications and hardware implementation as polynomials \cite{1560080}, leverage symbolic computer algebra techniques \cite{10.1145/3489517.3530605,8695790,8894250} 
% and hold the potential 
to solve verification problems.

\textbf{Multiplier Case Study}
Arithmetic circuits, especially those with large integer multipliers, play a key role in many areas. They are used in homomorphic encryption \cite{10.1007/978-3-642-41320-9_16}, security tasks such as malicious hardware detection \cite{6683016}, multimedia processing \cite{8105885}, and signal processing \cite{8105885}. We focus on multipliers because they belong to the challenging combinational arithmetic designs and showing our framework on multipliers suggests it is suitable for simpler blocks such as full adders \cite{8105885,wu2023gamora}.
However, verifying these circuits, particularly the large integer multipliers remains a challenge \cite{Kaufmann+2022+285+291,8894250,7167236}.
Large integer multipliers are also critical for tasks such as scientific computing \cite{bailey2002high}, financial operations \cite{hull2010high}, and cryptography \cite{paar2010understanding}. One approach uses Symbolic Computer Algebra (SCA) to verify different multipliers by recognizing full adders (\hw{FAs}) and half adders (\hw{HAs}) in the netlists. For example, \cite{7459464} shows that a 128-bit multiplier based on ripple carry adders takes more than one hour to verify, while a Booth multiplier takes over 100 hours when using the OneSpin commercial equivalence checker tool.

\textbf{GNN in EDA}
Graph neural networks (GNNs) learn graph structures and extract useful information \cite{kipf2016semi}. In EDA, circuit netlists naturally form graphs, which makes GNNs a good fit for this domain. For example, NeuroSAT \cite{DBLP:journals/corr/abs-1802-03685} uses message passing in a neural network to learn SAT problems and predict satisfiability. In another study \cite{8807085}, GNNs predict testability for netlists with performance similar to commercial tools. These ML and GNN techniques show promise for verification, but further work is needed to improve their performance and to address challenges in scalability and data management \cite{wu2023gamora}.

% We focus on two stages around technology mapping, as highlighted in Figure~\ref{fig: chip design}.
% One is before technology mapping, where we use the CSA multipliers and Booth multipliers as examples. The second one is post-technology mapping, and we will use 7nm mapped CSA multipliers and FPGA mapped CSA multipliers.

\section{GNN for Verification in Logic Synthesis}
The overview of GROOT framework is depicted in Figure \ref{fig:Overview_flow},
consisting of
% typically comprises 
five stages,
% In the remainder of this section, we provide more details
i.e., (a) Converting the netlist into a transitional graph representation using an open-source EDA tool ABC \cite{Mishchenko2007}; (b) Pre-process the transitional graph and generate the standardized logic synthesis-based EDA graph; (c) Partition of the large EDA graphs; (d) Utilize GNN for aggregation and message passing; and (e) Node classification and post-processing. 
\subsection{Converting Netlists into Transitional Graph}
A Boolean network (digital design) can be described as a directed acyclic graph (DAG), where the nodes symbolize logic gates and edges symbolize connecting wires. 
An And-Inverter Graph (AIG) represents a specific type of combinational Boolean network, comprised of two input \hw{AND} gates and \hw{inverters}  \cite{1688855}. Essentially, AIG graphs are specialized DAGs that encapsulate the logical functionality of Boolean networks.
Interestingly, through 
% the application of 
DeMorgan's rule, the combinational logic of any given Boolean network can be easily transformed into an AIG.

\begin{figure*}[t]
\centering
\includegraphics[width=0.87\linewidth]{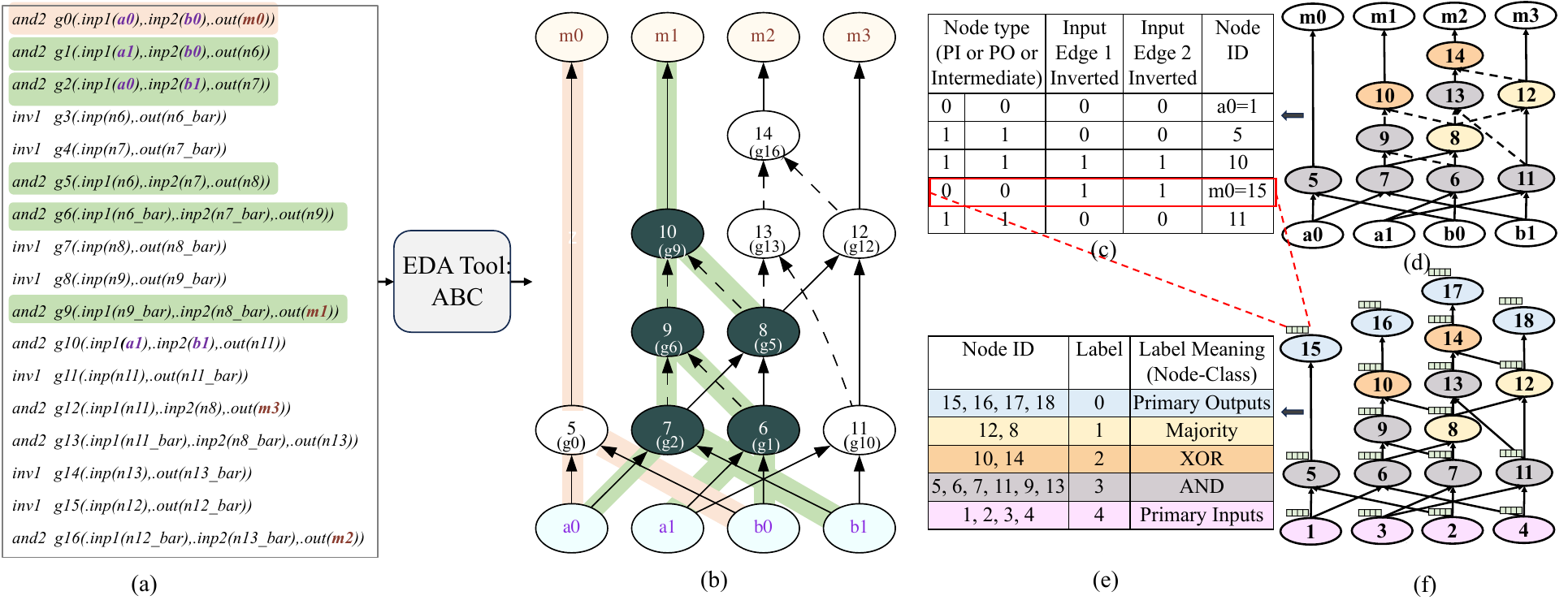}
 % \vspace{-12pt}
\caption{Input to EDA graph flow: (a) Two-bit multiplier netlist, (b) AIG representation of two-bit multiplier using ABC (the dotted line represents inverted inputs to node), (c) Node features (selected nodes shown), (d) EDA graph of two-bit CSA multiplier, (e) Ground truth labels for the GNN model. (f) EDA graph embedding with node features of two-bit CSA multiplier.}
\label{fig:transitinal graph}
\vspace{-10pt}
\end{figure*}

% In GROOT, the conversion process initiates with the post-synthesized netlist, transforming it into an AIG.
% And-Inverter Graph (AIG). 
In GROOT, this transformation is accomplished through an open-source EDA tool ABC \cite{Mishchenko2007}. Figure \ref{fig:transitinal graph} illustrates the transformation process using a two-bit CSA multiplier.
% as an example. 
The ABC takes a netlist as an input, as shown in Figure \ref{fig:transitinal graph} (a), and generates the corresponding AIG representation, shown in Figure \ref{fig:transitinal graph} (b).
% The ABC takes netlist as a input  (Figure \ref{fig:transitinal graph} (a) and create AIG representation of the corresponding netlis as shown in Figure \ref{fig:transitinal graph} (a).
% The ABC is equipped to process inputs in two formats: the netlist (Figure \ref{fig:transitinal graph} (a)), and the schematic (Figure \ref{fig:transitinal graph} (c)).
% In these figures, 
In AIG representation, inputs $a1a0$ and $b1b0$ represent the two-bit binary numbers for the multiplier and multiplicand, respectively. The multiplication result is represented using $m3m2m1m0$ bits.
For example, multiplier $a1a0=10$ and multiplicand  $b1b0=11$ gives the multiplication result $m3m2m1m0=0110$.
% This figure represents what we refer to as the transitional graph, detailing the progression from input to AIG graph. 
% ABC can accept input in two formats: either the netlist shown in Figure \ref{fig:transitinal graph} (a), or the schematic shown in Figure \ref{fig:transitinal graph} (c). Either input format allows ABC to generate an AIG graph.
% The multiplier and multiplicand are two bit binary numbers. In figures  \ref{fig:transitinal graph} (a) and \ref{fig:transitinal graph} (c) $a1a0$ represents two bits of multiplier and $b1b0$ represents two bits of multiplicand.
% \textcolor{red}{Describe the figure a, b,c in the figure}
% \textcolor{red}{describe,input, output and function, also use the example to describe the multiplication}
% The resulting AIG graph output from ABC is depicted in Figure \ref{fig:transitinal graph} (b).
% Within this graph, 
The multiplication of the least significant bits (LSB) highlighted in golden color, symbolizing the \hw{`AND'} operation at node 5 (i.e., $m0=a0\cdot b0$). 
The additional operation of multiplication is \hw{`XOR'} (green), performed at  nodes 6, 7, 8, 9, and 10, and can be represented by the equation 
$m1=a0\cdot b1$ $xor$ $a1\cdot b0$. The \hw{`NOT'} operations
% within the AIG representation
are indicated by dashed lines.

\subsection {Node Features and  Node Label creation }
We take the node and edge information from the AIG representation (transitional graph) to form the EDA graph. 
% We define a specific graph that focuses on circuit design.
% , referred to as a circuit design-based EDA graph.
% We define, 
We define circuit-based EDA graph as ${G = (V,E)}$ with ${N}$ nodes $v_i \in {V}$ and edges $(v_i,v_j) \in {E}$. We use an adjacency matrix $A \in R^{N \times N}$ to 
 % visualize
 describe graph connections,
 % of the graph, 
 a degree matrix ${D}_{ii} = \sum_{j} A_{ij}$ and a feature matrix $X=\lbrace {x}_{1},{x}_{1},...,{x}_{N} \rbrace$.

We transform an AIG graphs (Figure \ref{fig:transitinal graph}, (d)) into graph embedding. We create node features by explicitly encoding two aspects: node type and input edge polarity. Each node in the AIG belongs to one of three categories Primary Inputs (\hw{PI}), Internal logic gates (\hw{AND} nodes), or Primary Outputs (\hw{PO}) and is represented by a 4-bit vector as shown in figure \ref{fig:transitinal graph}, (c).

First two bits (node type) are encoded as:`00' for Primary Input (\hw{PI}), 
`11' for Internal logic node (\hw{AND} gate), `0X' for  Primary Output (\hw{PO})  where `X' indicates polarity inherited from the preceding internal node. 
Last two bits (input polarity for internal nodes) are encoded as:
`00' for Both inputs non-inverted 
`01' or `10' for Exactly one input inverted (`01' for right, '`10' for left input) 
`11' for  Both inputs inverted.
\hw{PI} nodes have no incoming edges, thus their polarity bits are always '00'. \hw{PO} nodes inherit polarity information from preceding internal nodes. 
For instance (Figure \ref{fig:transitinal graph}, (c) vector table), node 5, an internal node with non-inverted input edges, has a feature vector of 1100. Similarly, node 10, another internal node with inverted inputs, has a feature vector of 1111. The \hw{PI} (node 1) or $a0$ has a feature vector of 0000 since no input connections, while the \hw{PO} node 15 or $m0$ has a feature vector of 0011  since it inherited non inverted input as highlighted in red dotted lines between the figures \ref{fig:transitinal graph} (c) and \ref{fig:transitinal graph} (f). Our EDA graph embedding consist of four-node features, a distinction from the three-node features in GAMORA~\cite{wu2023gamora}. Implementing additional node features
% offers benefits compared to GAMORA~\cite{wu2023gamora}, including 
offers a more robust representation of nodes and improved generalization.
% With the addition of more features, 
% Our model possesses the capacity to learn from a broader spectrum of circuit design characteristics.

Next, we create labels for the ground truth using ABC \cite{Mishchenko2007}. Figure \ref{fig:transitinal graph} (c), depicts the labels for the two-bit CSA multiplier. For nodes 1 to 4 (\hw{PI} nodes), we label them as 4. For nodes 5, 6, 7, 9, 11, 13 (two-input \hw{AND} gates), we label as 3. For nodes 10 and 14 (\hw{XOR}), we label as 2. For nodes 12 and 8 (\hw{MAJ} functionality) are labeled as 1. Lastly, all \hw{PO} nodes, namely 15 to 18, are labeled as 0. 
\subsection{GNN and Edge Re-growth After Partition}
To handle the memory footprint of large EDA graphs, we partition the graph into subgraphs as shown in Figure~\ref{fig:Overview_flow} (c) and feed them to a GNN for node classification. We use GraphSAGE~\cite{https://doi.org/10.48550/arxiv.1706.02216}. The layer takes a graph with node features as input (Figure~\ref{fig:Overview_flow} (d)) and aggregates features. We then apply METIS~\cite{karypis1998fast} to partition the graph.

Let $p$ index a partition and let $S_p$ be its node set. For a node $u$, $\mathcal{N}(u)$ is the set of nodes one hop from $u$. For edges, $(i,j)\in E$ denotes a directed edge from node $i$ to node $j$, and $E[S]$ denotes the edges whose endpoints are both in $S$.

For partition $p$, we collect one-hop neighbors and the boundary nodes:
\begin{equation}
N(S_p)=\bigcup_{u\in S_p}\mathcal{N}(u),\qquad
B_p = N(S_p)\setminus S_p,
\label{eq:Np_def}
\end{equation}
where $N(S_p)$ is the set of all nodes that are exactly one hop away from any node in $S_p$, and $B_p$ keeps only those neighbors that are outside $S_p$ (the boundary nodes of partition $p$).

Next, we define the edges that cross between $S_p$ and its boundary $B_p$, and the augmented sets:
\begin{equation}
\begin{aligned}
C_p &= \{\, (i,j)\in E \mid (i\in S_p \wedge j\in B_p)\ \vee\ (i\in B_p \wedge j\in S_p) \,\},\\
S_p^{+} &= S_p\cup B_p,\\
E_p^{+} &= E[S_p]\cup C_p.
\end{aligned}
\label{eq:CpVp_def}
\end{equation}
Here, $C_p$ contains all edges with one endpoint in $S_p$ and the other in $B_p$; $S_p^{+}$ and $E_p^{+}$ are the node and edge sets after adding these boundary nodes and crossing edges.
\begin{algorithm}
\caption{Graph Boundary Edge Re-growth}
\label{alg:boundary_alternative}
\begin{algorithmic}[1]
\REQUIRE $G=(V,E)$; partitions $\{S_p\}_{p=0}^{n}$ 
\FOR{$p = 0$ \textbf{to} $n$}
    \STATE $B_p \leftarrow \Bigl(\bigcup_{u\in S_p}\mathcal{N}(u)\Bigr)\setminus S_p$ \hfill \COMMENT{boundary nodes; Eq.~\eqref{eq:Np_def}}
    \STATE $C_p \leftarrow \{(i,j)\in E : (i\in S_p \wedge j\in B_p)\ \vee\ (i\in B_p \wedge j\in S_p)\}$ \hfill \COMMENT{crossing edges; Eq.~\eqref{eq:CpVp_def}}
    \STATE $S_p^{+} \leftarrow S_p \cup B_p$; \quad $E_p^{+} \leftarrow E[S_p] \cup C_p$ \hfill \COMMENT{augmented sets; Eq.~\eqref{eq:CpVp_def}}
\ENDFOR
\RETURN $\{(S_p^{+}, E_p^{+})\}_{p=0}^{n}$
\end{algorithmic}
\end{algorithm}
Algorithm~\ref{alg:boundary_alternative} performs boundary re-growth per partition: it first identifies the boundary nodes $B_p$ using Eq.~\eqref{eq:Np_def}, then collects all crossing edges $C_p$ using Eq.~\eqref{eq:CpVp_def}, and finally forms the augmented node and edge sets.

We observe that EDA graphs contain approximately $10\%$ boundary edges between partitions, and the boundary recovery process does not add significant complexity to the inference stage. This approach regenerates boundary edges between disconnected partitions to prevent the loss of features and supports message passing between inter-partition nodes.
\subsection{Node Classification and Post-Processing }
We use GNN to classify the nodes into two categories \hw{XOR} and \hw{MAJ} as depicted in Figure \ref{fig:Overview_flow} (e). We use the algebraic re-writing technique \cite{8105885, 8695790} for verification. 
The algebraic representation of the basic Boolean operators is summarized in the Table \ref{table:boolean-operators}.
\begin{table}[h]
\centering
\caption{Algebraic Representations of Basic Boolean Operators ($a, b, c$ are inputs)}
\label{table:boolean-operators}
\begin{tabular}{|l|l|l|}
\hline
\textbf{Operation} & \textbf{Boolean Model} & \textbf{Algebraic Model} \\
\hline
NOT & $\neg a$ & $1 - a$ \\
\hline
AND & $a \land b$ & $ab$ \\
\hline
XOR & $a \oplus b$ & $a + b - 2ab$ \\
\hline
XOR3 & $a \oplus b \oplus c$ & $a + b + c - 2ab - $\\ & & $ 2ac - 2bc+ 4abc$ \\
\hline
MAJ & $(a \lor b) \land (a \lor c)$ & $ab + ac + bc - 2abc$ \\
\hline
\end{tabular}
\end{table}
Consider the case for \hw{XOR3} and \hw{MAJ} operations. The sub-polynomial expression is   $x_1 + 2x_2 + \ldots$ , where $x_1 = \hw{XOR3}(a, b, c)$ and $x_2 = \hw{MAJ}(a, b, c)$, where $a, b, c$ 
are inputs of \hw{XOR3} and \hw{MAJ} functions. Substituting the algebraic representations of \hw{XOR3} and \hw{MAJ} into the sub-polynomial, we obtain:
\begin{align*}
x_1 + 2x_2 + \ldots &= (a + b + c - 2ab -2ac -2bc + 4abc) \\
&\quad + 2(ab + ac + bc - 2abc) \\
&= a + b + c
\end{align*}
This simplification results in the elimination of the four nonlinear terms: $2ab$, $2ac$, $2bc$, and $4abc$.

This polynomial reduction based on algebraic re-writing \cite{8105885, 8695790} and is integrated in ABC \cite{Mishchenko2007}. This approach is reliant on detecting \hw{XOR} and \hw{MAJ} gates from a flattened netlist, a process that tends to be time-consuming.  We leverage the GNN node classification to detect \hw{XOR} and \hw{MAJ} gates which makes verification efficient.  As depicted in Figure \ref{fig:Overview_flow} (e), in the two-bit CSA multiplier, nodes 10 and 14 are classified as \hw{XOR}, while nodes 12 and 8 are classified as \hw{MAJ}. 
These nodes are used for verification using the method described in \cite{8105885}. Note that the accuracy of node classification directly translates to the verification accuracy.
%These nodes are subsequently used for verification with the methodology described in \cite{8105885}. Note that the node classification accuracy directly translate to the verification acuuracy.
In the following section, we describe kernel design which is necessary to train GNN with larger multipliers. More details are in subsection \ref{sec:accuracy} Accuracy Enhancement with Large Design Training part.

\section{Kernel Design - GROOT-GPU}

\begin{figure}[t]
 \centering
   \includegraphics[width=.85\linewidth]{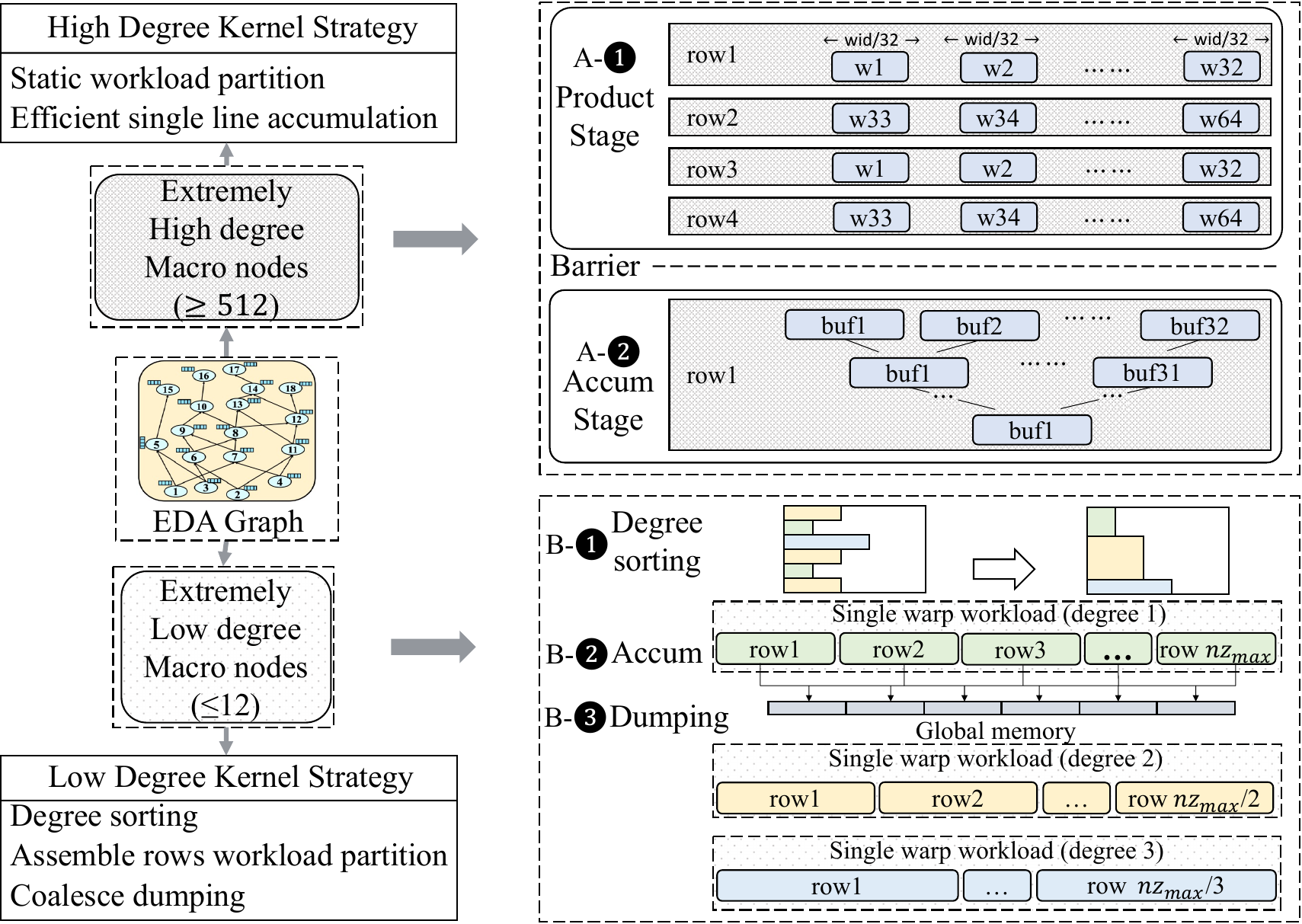} 
    \caption{GPU Kernel Design for EDA.}
     \label{fig:eda_kernel_design_overview}
     % \vspace{-15pt}
%\includepdf[pages=-]{Doc1-cropped.pdf}
\end{figure}

\begin{figure*}
\centering
\includegraphics[width=.85\textwidth]{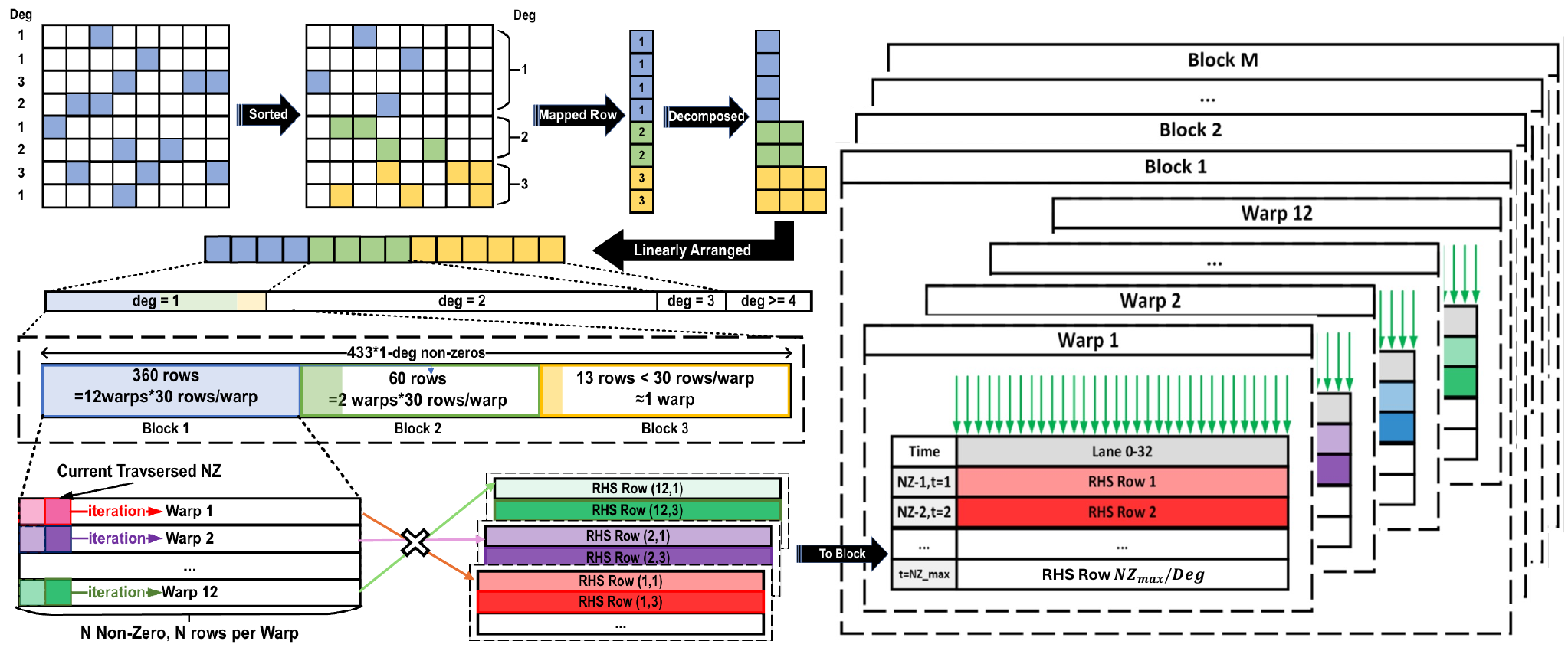}
% \vspace{-10pt}
    \caption{ Detailed process of LD-kernel, from degree-sorting, row-assembling, block-partitioning, to warp-wise multiplication and summation, with block-wise parallelism.}
    \label{fig:Warp-Block-Op}
\end{figure*}

We tailor GPU kernels (high-degree (HD) kernel and low-degree (LD) kernel) separately for the extremely high-degree macro nodes ($\ge512$) and the low-degree macro nodes ($\leq12$).
% (EDA graph observation please see Appendix \ref{Appendix:kernel}). 
The whole GPU kernel is programmed in CUDA C. 
% The codebase will be released with the paper.
We start by partitioning the workload (non-zero elements) statically for each row of the adjacency ($A$) matrix (all nodes possessing a degree equal to the width). 
We show an example in the Fig. \ref{fig:eda_kernel_design_overview}. 
% or example, 
The HD macro nodes contain 4 rows, namely $row1$ to $row4$. 
Each row contains $wid$ non-zero elements. Each block in the kernel contains 64 warps, numbered from $w1$ to $w64$. 
We divide each row into 32 equal workloads, each containing $\frac{wid}{32}$ non-zero elements. Then we assign the workloads in $row1$ to the warps numbered $w1$ to $w32$ in turn, and assign the workloads in $row2$ to warps from $w33$ to $w64$. 
The LD-kernel design for low-degree macro nodes is shown in the lower half of Fig. \ref{fig:eda_kernel_design_overview}. Step B processes the degree sorting on the adjacent matrix with the following steps:
(1) computing each row’s degree using the row pointer array with time complexity of $\mathcal{O}(n)$ when employing count sort \cite{sun2009count} or radix sort \cite{merrill2011high}, with $n$ indicating the number of rows; (2) applies a stable sorting algorithm to sort rows by their degrees; and (3) updating the row pointer array to reflect the new rows’ order, with time complexity of $\mathcal{O}(n)$. The dominant time complexity of this operation arises from applying the stable sorting algorithm. Nevertheless, employing count sort, a linear time-complexity algorithm, can optimize the overall time complexity to $\mathcal{O}(n)$. The details of sorting and row-assembling are as described in Figure ~\ref{fig:Warp-Block-Op}. The warp-block operation of LD-kernel starts with sorting on the original sparse input by the degree of each row. It maps the rows into an array linearly, in which the partitioning is executed by dividing the array into sections of rows according to their degree, sequentially from the smallest to the largest. Then, rows with the same degree in every section are further partitioned into blocks, whose warps will be processed in parallel to extract rows and multiply them with the corresponding right-hand-side (RHS) rows from the dense input. The resulting product rows sum up by the degree of the left-hand side rows to produce the output rows. This blockwise operation is performed in parallel with other blocks. For example, in the lower left part of the figure, warp 1 traverses its non-zeros one by one from the left to the right, while locating the corresponding RHS row (1,1). The first 1 implies the warp 1, the second 1 refers to the first row, which warp 1 is responsible for. Then, the first traversed non-zero from warp 1 multiplies the RHS row (1,1), resulting in the output RHS row 1 in the right part of the figure, and so forth for the rest of warp 1 and other warps. Multiple rows of non-zero elements are assigned to the same warp, rather than the maximum number of non-zero elements from one row per warp. Since the degree of each row is small, say 3, the number of warps per block in this kernel is set to $6m$. For rows with degrees of 1, 2, and 3, each warp is responsible for $6m$, $3m$, and $2m$ rows, which translates into $nz_{max}$, $nz_{max}/2$ and $nz_{max}/3$ rows in the figure, respectively. This achieves workload balance between warps and helps to increase the proportion of active warps.  By aggregating the workload of many small degree rows into the same warp, we will significantly reduce the overhead of warp switching and improve calculation efficiency compared to the ordinary SPMM algorithm process. Moreover, this aggregated organization method will further enhance access continuity to global memory when transferring calculation results from shared memory to global memory. We can utilize the technique of coalesce dumping to exploit this advantage fully.
During step $B-\ding{3}$, termed coalesce dumping, we will write the intermediate results of multiple consecutive rows, handled by the same warp, into a continuous area in global memory. This approach will significantly enhance the efficiency of accessing global memory.

\section{Performance Evaluation}

% \section{Performance Evaluation}
This section presents the evaluation of GROOT by comparing its memory usage, and run-time against two baselines: the traditional open source tool ABC \cite{Mishchenko2007} and the GNN-based SOTA GAMORA~\cite{wu2023gamora}. When comparing with \cite{wu2023gamora}, we run the  GAMORA framework against our modified datasets. We use a Linux-based host with AMD EPYC 7543 32-Core Processor  and an NVIDIA A100-SXM 80 GB.

\subsection{Accuracy and Accuracy Recovery of Various Multipliers}
\label{sec:accuracy}
\begin{figure*}[t]
 \centering
\includegraphics[width=\linewidth]
 {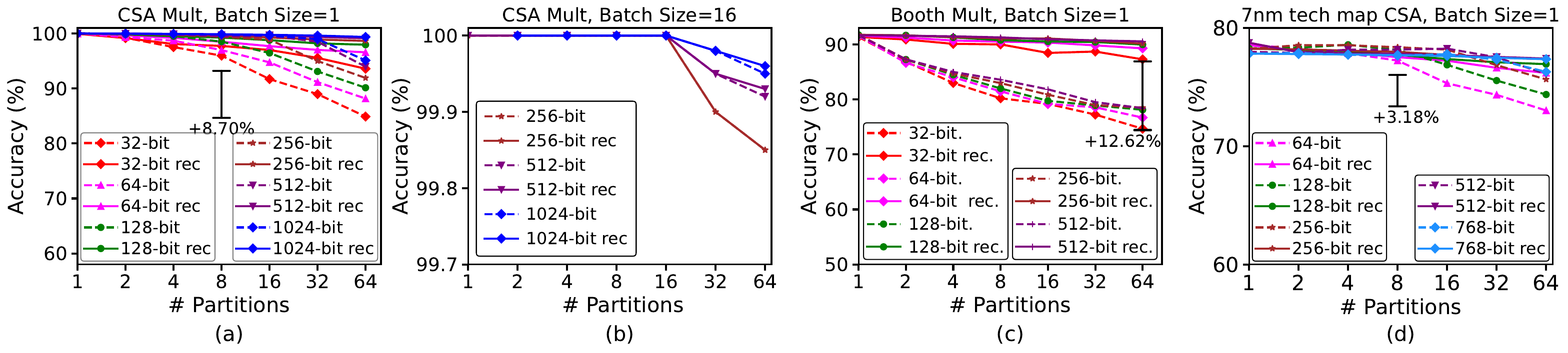}

 \caption{Verification accuracy as a function number of partitions for (a, b) CSA multipliers, batch size 1 and 16, respectively; (c) Booth multipliers, batch size 1; (d) CSA multipliers after 7nm technology mapping, batch size 1. All the multipliers were trained using 8-bits.}
     \label{fig:accuracy_combine}
\end{figure*}

% \noindent\textbf{Accuracy Results.}
Our GNN model is trained on an 8-bit multiplier and then used in inference on larger multipliers of the same dataset. Figure \ref{fig:accuracy_combine} depicts the accuracy with respect to the number of partitions for different datasets. Figure \ref{fig:accuracy_combine} (a) shows the accuracy on CSA multipliers with batch size one. Without any partitioning, we achieve high accuracy, reaching 100\% for multipliers of sizes 128-bits and above, while the accuracy is 99.94\% for the 32-bit multiplier.
% Introducing partitions leads to a marginal accuracy loss. 
% As the partition count grows, this accuracy loss becomes more pronounced, echoing the reality that more partitions necessitate the removal of more boundary edges. 
As the number of partitions increases, the loss in accuracy becomes more noticeable because more partitions require the removal of more boundary edges.
% This effect is particularly noticeable in smaller multipliers, where each removed edge plays a substantial role. For instance, 
% Figure \ref{fig:memory_combine} (a) reveals that partitioning a 32-bit multiplier into 64 subgraphs causes its accuracy to dip to 84.89\% (represented by the red dashed line). This drop results from removing a substantial portion of the total edges.
% This drop can be attributed to the significant proportion of edges removed in relation to the total edge count.
However, using our boundary edge re-growth approach effectively recovers accuracy.
% On the other hand, leveraging our boundary edge re-growth strategy effectively recovers accuracy.
% counters this decline in accuracy. 
In Figure \ref{fig:accuracy_combine} (a), the solid line denotes the regained accuracy when using boundary edge re-growth.
% , while the dashed line represents the accuracy prior to this recovery technique. 
Notably, this edge re-growth method achieves a maximum recovery of 8.7\% boost in accuracy of a 32-bit multiplier. 
% A vital enhancement, especially for the 32-bit multiplier that witnesses the most significant degradation due to graph partitioning. 
% In general, as the number of partitions increases, this recovery becomes increasingly evident. 
By adopting our edge re-growth approach, one can afford to use more partitions while maintaining high accuracy
% \subsubsection{Scalability and Design complexity}
\label{subsec:design_complexity}
To evaluate the scalability of GROOT, we evaluate its accuracy on  large CSA multipliers such as the 1024-bit multiplier with a batch size of 16 containing 134,103,040
nodes and 268,140,544 edges. The figure \ref{fig:accuracy_combine} (b) shows accuracy with respect to number of partions. 
The trends show that the accuracy is at 100\% up until 16 partitions. This accuracy can be attributed to the presence of a large number of edges in these large graphs and a small number of edges removal does not affect message passing in GNN. 
Consequently, Partitioning does not much impact the accuracy. 
However, post the 16-partition mark, there is a slight drop in accuracy since more edges are removed to create partitions. 

To evaluate GROOT's performance with complex graphs, we utilize the Booth multipliers dataset. 
% , with their intricate structures. 
Figure \ref{fig:accuracy_combine}  (c)
shows accuracy with respect to the number of
partitions.  The accuracy drop is more compared to that in other datasets. However, the utilization of the edge re-growth approach enables the mitigation of this accuracy drop, as illustrated by the solid line in Figure \ref{fig:accuracy_combine} (c). The re-growth achieves a maximum 12.62\% accuracy recovery in a 32-bit multiplier.
Additionally, we test with the ASAP 7nm technology \cite{xu2018standard} mapped netlist dataset. This netlist comprises 161 standard cell gates, including the multi-output gate, leading to certain irregularities. As evidenced in Figure \ref{fig:accuracy_combine}  (d), even with such irregularity, GROOT shows high accuracy and maintains more than 76\% accuracy after edges re-growth. In summation, GROOT is capable of handling design complexities.
% \subsubsection{Accuracy Improvement with Large Design Training}

\begin{figure}[t]
 \centering
\includegraphics[width=\linewidth]
 {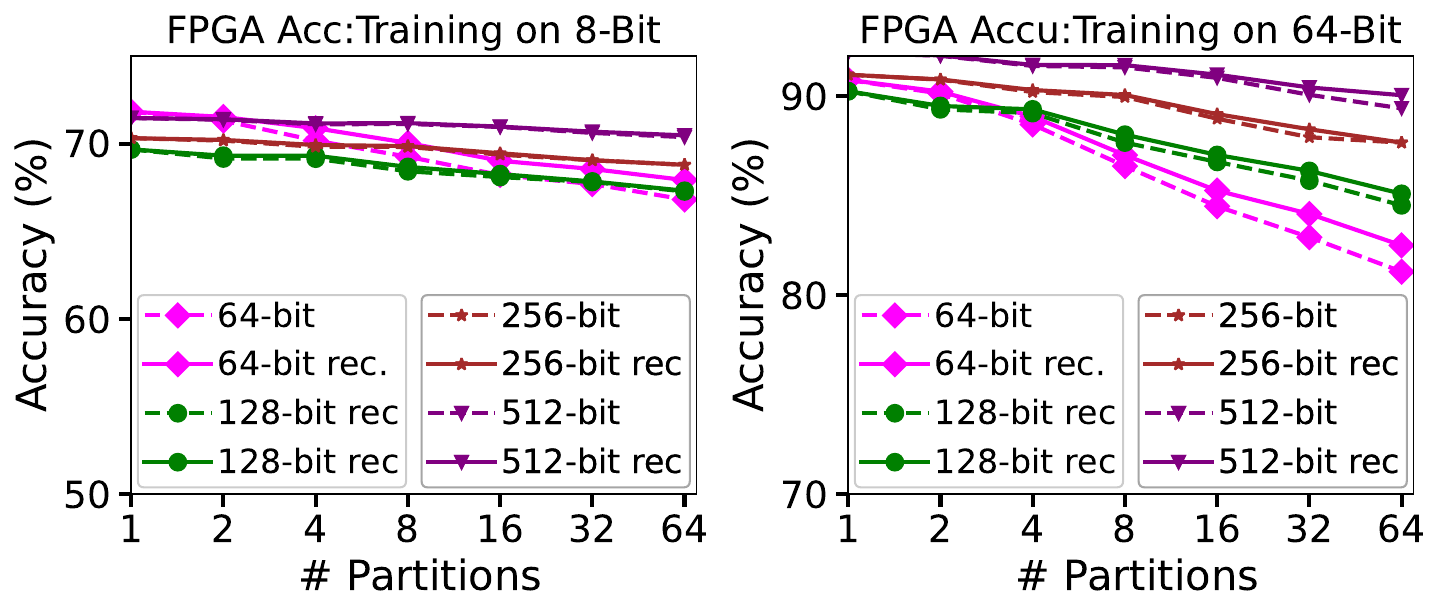}
  \caption{FPGA multiplier dataset accuracy improvement (a) 8-bit training, (b) 64-bit training}
     \label{fig:fpga_results}
     \vspace{-10pt}
\end{figure}

Figure \ref{fig:fpga_results} (a) shows the accuracy of FPGA-mapped CSA multipliers with batch size equal to 1 and the model is trained on 8 bits. The accuracy is low among all the datasets.
To further improve prediction accuracy, we focus on training the GNN model using larger multipliers.
% to better integrate these complexities. 
This approach significantly improves the model's prediction accuracy.
% , as shown in Figure \ref{fig:accuracy_improvement}. For FPGA mapped multipliers, 
When the model trained on a 64-bit multiplier boosts the accuracy for a 64-bit multiplier from 71.82\% (Figure \ref{fig:fpga_results} (a), number of partition=1) to 90.8\% (Figure \ref{fig:fpga_results} (b), number of partition=1) an 18.98\% boost in accuracy.
% Similarly, in 7nm technology mapping, training on a 128-bit mapped multiplier elevates 64-bit multiplier accuracy from 78.55\% to 89.23\%, a 10.68\% increase which can be observed in Figure \ref{fig:accuracy_improvement} (b).
However, this accuracy gain comes with increased training time. Training a 64-bit FPGA for 100 epochs takes 2914.42 seconds. To mitigate this time cost, we propose designing a specialized kernel for faster matrix multiplication, a major factor in training and inference time.

\subsection{Memory Footprint Analysis}
\label{sec:memory}
\begin{figure*}[t]
 \centering
\includegraphics[width=0.98\linewidth]
 {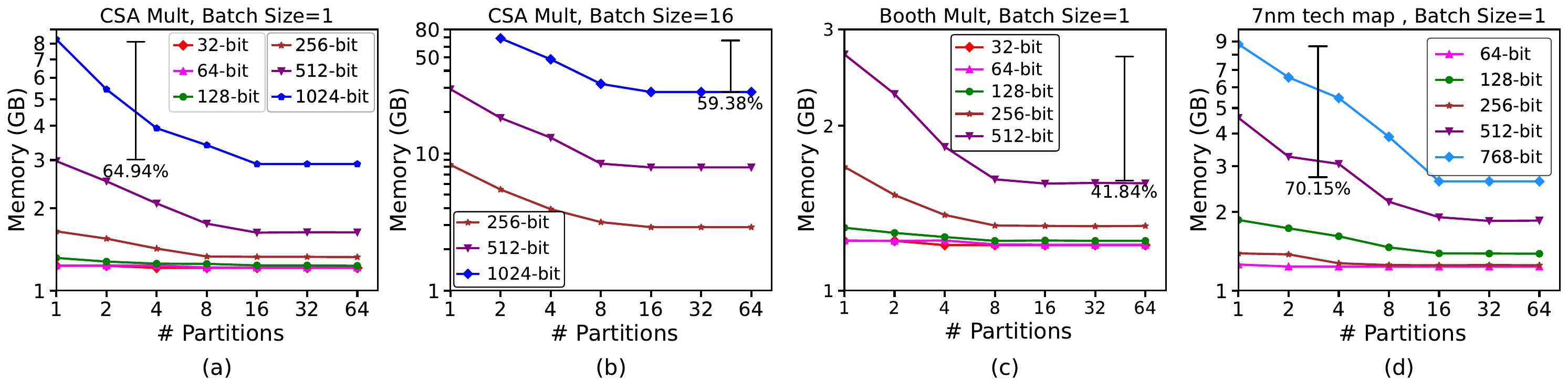}
 \caption{Memory utilization as a function of the number of partitions for (a) CSA multipliers, for a batch size of 1, (b) CSA multipliers, for a batch size of 16, (c) Booth multipliers, for a batch size of 1, and (d) CSA multipliers, following the application of 7nm technology mapping, with a batch size of 1.}
    % \caption{Comparative analysis of memory usage across various multipliers trained on an 8-bit model: (a) Memory utilization vs the number of partitions CSA multipliers with batch size=1. (b) Memory utilization vs the number of partitions for CSA multipliers with batch size=16. (c)Memory utilization vs the number of partitions for Booth multipliers with batch size=1. (d) Memory utilization vs the number of partitions for CSA multipliers after 7nm technology mapping with batch size=1}
     \label{fig:memory_combine} 
     % \vspace{-12pt}   
\end{figure*}
% \begin{figure*}[ht]
%  \centering
%    \includegraphics[width=\linewidth]{Figures/acceleration ratio v3_cropped.pdf} 
%  %\includegraphics[scale=0.6]{Figures/Accuracy_vs_partions_new.pdf}
%  % \vspace{-18pt}
%     \caption{The runtime comparison among GROOT-GPU and SOTA GPU-based GPU Kernel designs, where the acceleration ratio of 1 from GNNAdvisor is drawn as the black dash line.}
%      \label{fig:eda_kernel_design}
%      % \vspace{-0.1in}
% %\includepdf[pages=-]{Doc1-cropped.pdf}
% \end{figure*}

% \begin{figure}[t]
%  \centering
% \includegraphics[width=.95\linewidth]
%  {Figures/memory_vs_partitions_sca_nvidia_smi_large.pdf}
%     \caption{Memory utilization of CSA multipliers (batch size=16) with graph partitioning.}
%      \label{fig:memory_large}
% \end{figure}
\begin{table}[ht]
\begin{center}
\small
 \scriptsize
\caption{\label{table-memory} Large Multiplier GPU Memory Usage (In MB) Comparison. (Batch size=16,  OOM= Out of Memory).}
\resizebox{0.9\columnwidth}{!}{
\begin{tabular}{|c| c | c |c |c |c|} 
 \hline 
 \textbf {\# Part.}  &\textbf{ 256-Bit} & \textbf{ 512 -Bit} &  \textbf{ 1,024-Bit } \\
 % [0.3ex]
 \hline \hline
 GAMORA~\cite{wu2023gamora} & 8,263 &  29,375 & \textbf{OOM}\\ 
 \hline
 GROOT 2 Part.  & 5,457 & 18,135 & {68,923} \\ 
 \hline
 GROOT 4 Part.  & 3,923 & 13,025 & {48,463} \\
 \hline
 GROOT  8 Part. & 3,157 & 8,421 & 32,093 \\
 \hline
 GROOT 16 Part.  & 2,901 &  7,909 & 27,997 \\
 \hline
 GROOT 32 Part. & 2,901 & 7,909 & 27,997  \\
 \hline 
 GROOT 64 Part.  & 2,901 & 7,909& 27,997  \\
 \hline
\end{tabular}}
% \vspace{-0.1in}
\end{center}
\end{table}
Figure \ref{fig:memory_combine} illustrates the GPU memory utilization by various multipliers with respect to the number of partitions. Figure \ref{fig:memory_combine} (a) shows the GPU memory utilization on the y-axis and the number of partitions on the x-axis for CSA multipliers with batch size one. As the number of partitions increases, the memory requirement decreases. For larger multipliers (e.g., 1024 bits), the memory reduction trend follows an exponential decay. When partitioned into 64 sub-graphs, the 1,024-bit 
multiplier showed a maximum benefit of 64.94\% reduction in memory requirement as per depicted in the Figure \ref{fig:memory_combine} (a).

To evaluate the scalability of GROOT, we evaluate its performance on massive multiplier graphs such as the 1024-bit multiplier with a batch size of 16 containing 134,103,040 nodes and 268,140,544 edges, as depicted in Figure \ref{fig:memory_combine} (b). Partitioning the 1,024-bit multiplier into 64 sub-graphs resulted in a maximum memory reduction of 59.38\%. Without partitioning, even high-end GPUs such as NVIDIA A100-SXM with 80 GB memory cannot perform verification on this massive graph. Thus, GROOT offers a fundamental solution to scalability. 
Our method is different from GAMORA \cite{wu2023gamora}, which requires multiple GPUs to handle massive graphs while we only need one low-end GPU.
Table \ref{table-memory} shows the different multipliers and their GPU memory utilization.

Furthermore, to recover the accuracy, our algorithm regrows the edges after partitioning. The effect of the number of partitions on the memory requirement can be observed until the number of partitions is equal to 16. When the partitioning size is large (say 32) as shown in Figure \ref{fig:memory_combine} (b), the recovered edge consumes a large portion of the memory footprint, thus we observe less memory saving.

To demonstrate GROOT's effectiveness on complex designs, we evaluate it on different complex datasets. Figure \ref{fig:memory_combine} (c) shows memory utilization versus the number of partitions for booth multipliers, indicating an exponential reduction in memory with respect to partitions. The 512-bit booth multiplier shows maximum memory requirement reduction which is 41.84\%. Figure \ref{fig:memory_combine} (d), shows memory utilization versus number partitions for 7nm mapped CSA multipliers, demonstrating a significant reduction in memory requirement for post-technology-mapped CSA multipliers. For instance, the maximum memory requirement reduction for the 768-bit multiplier is 70.15\%. Similarly,  Figure \ref{fig:fpga_results} (c) shows memory utilization for FPGA-mapped CSA multipliers. The maximum memory requirement reduction is 57.62\%  for a 512-bit multiplier. 
The benefits of memory reduction remain with increased design complexity. 
% To demonstrate GROOT's effectiveness on complex designs, we evaluate it on various complex datasets. Figure~\ref{fig:memory_combine}~(c) shows memory utilization versus the number of partitions for Booth multipliers, highlighting an exponential reduction in memory with respect to partitions. The 512-bit Booth multiplier achieves a maximum memory reduction of 41.84\%.
% Figure~\ref{fig:memory_combine}~(d) shows significant memory reduction for post-technology-mapped CSA multipliers. For example, the 768-bit multiplier achieves a maximum memory reduction of 70.15\%.

% \begin{figure*}[ht]
%  \centering
%    \includegraphics[width=\linewidth]{Figures/acceleration ratio v3_cropped.pdf} 

%     \caption{The runtime comparison among GROOT-GPU and SOTA GPU-based GPU Kernel designs, where the acceleration ratio of 1 from GNNAdvisor is drawn as the black dash line.}
%      \label{fig:eda_kernel_design}
% \end{figure*}
\begin{figure*}[ht]
 \centering
   \includegraphics[width=\linewidth]{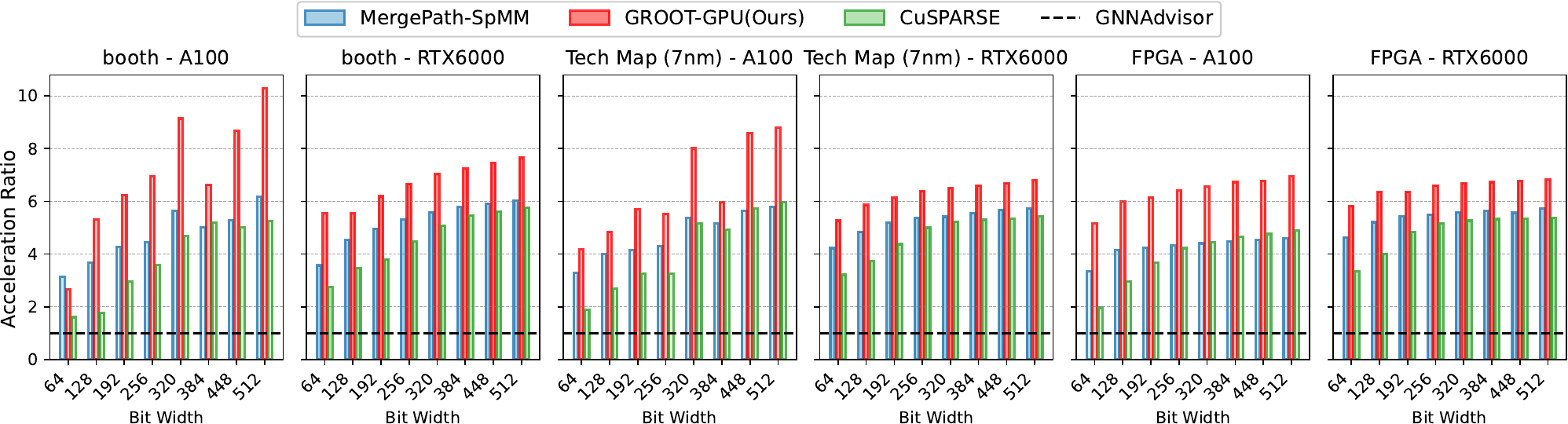} 

    \caption{The runtime comparison among GROOT-GPU and SOTA GPU-based GPU Kernel designs, where the acceleration ratio of 1 from GNNAdvisor is drawn as the black dash line.}
     \label{fig:eda_kernel_design}
     % \vspace{-10pt}
\end{figure*}
\subsection{Run Time Analysis and Comparative Study}
\begin{figure}[ht]
 \centering
 \includegraphics[width=.85\linewidth]{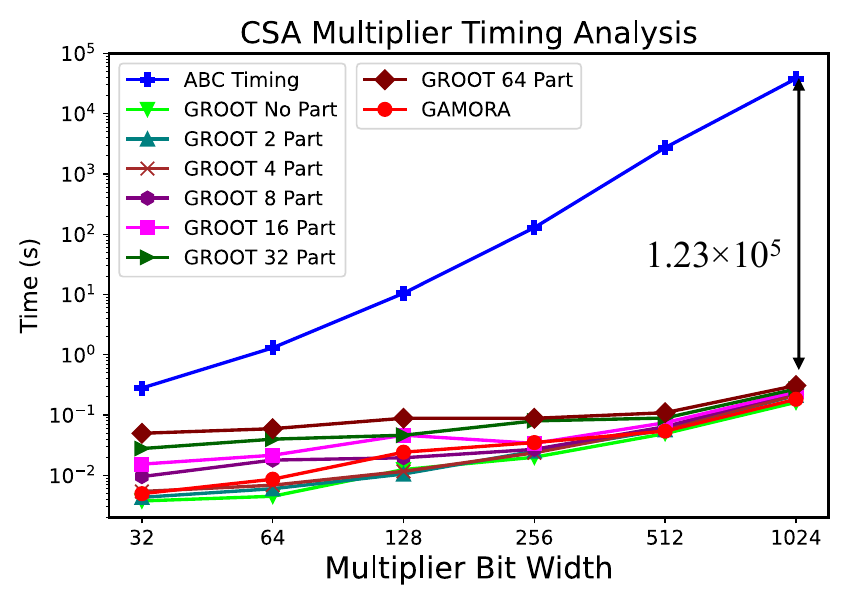}
    \caption{CSA multiplier verification time comparisons with ABC \cite{Mishchenko2007} and GAMORA \cite{wu2023gamora}.}
     \label{fig: timing improvement}
     \vspace{-15pt}
\end{figure}

Figure \ref{fig: timing improvement} shows the inference run time of verifying different multipliers of different widths after applying boundary edge re-growth for accuracy recovery.
% Additionally, a comparative analysis is conducted between GROOT and two leading-edge methods.
As shown in Figure \ref{fig: timing improvement} (a), as the bit width increases, ABC's run time expands exponentially compared to both GROOT and GAMORA. 
% It is also important to recognize that the run time for GROOT depends upon the number of partitions: an increase in the number of partitions slightly increases the run time. 
% Nonetheless, this increment in subgraphs has the advantageous side effect of trimming down the memory demands, thereby facilitating the processing of more sizable graphs.
% % In a direct comparison, 
% In comparison, GROOT significantly outperforms ABC~\cite{Mishchenko2007}. 
% % To offer a specific instance, 
% When processing graphs for 1,024-bit CSA multipliers partitioned into 64 subgraphs, GROOT achieves a speedup of $1.23 \times10^{5}$ over ABC. Furthermore, the verification times exhibited by different partitioned graphs using GROOT align closely with GAMORA~\cite{wu2023gamora}. It is important to recognize that the verification time for GROOT depends upon the number of partitions: an increase in the number of partitions slightly increases the verification time due to a small partition time. 
% % It's pivotal to highlight, however, that 
% It is important to highlight that neither GAMORA nor ABC can efficiently handle large graph datasets on a single GPU, 
% % even one as advanced as the A100-SXM boasting 80GB memory, particularly when the batch size is adjusted to 16, 
% as depicted in Figure~\ref{fig:memory_edited}.
In comparison, GROOT significantly outperforms ABC~\cite{Mishchenko2007}. For instance, when processing graphs for 1,024-bit CSA multipliers partitioned into 64 subgraphs, GROOT achieves a speedup of $1.23 \times 10^{5}$ over ABC. Additionally, the verification times for partitioned graphs using GROOT closely align with GAMORA~\cite{wu2023gamora}. 
It is important to note that the verification time for GROOT depends on the number of partitions, with more partitions slightly increasing the verification time due to additional partitioning overhead. However, neither GAMORA nor ABC can efficiently handle large graph datasets on a single GPU, as shown in Figure~\ref{fig:memory_edited}.
\subsection{GPU kernel results}
We compare our design with SOTA GPU-based GPU Kernel designs such as 
% will be illustrated in Fig. \ref{fig:eda_kernel_speedup}, comparing them to 
cuSPARSE \cite{naumov2010cusparse}, MergePath-SpMM~\cite{shan2023mergepath},
% the academical frameworks such as 
GNNAdvisor~\cite{wang2021gnnadvisor}.
% \textbf{Add new section GPU kernel design}
% \textcolor{red}{direct description and indirect description}
% The Figure~\ref{fig:eda_kernel_design} shows the overall evaluation results, in which MergePath--SpMM, our GROOT--GPU, and CuSPARSE are in comparison of SpMM runtime in acceleration ratio against GNNAdvisor, which is present as the black horizontal dash line indicating the 100\% acceleration ratio. The kernels are tested on Booth Multiplier, Technology Mapping, and FPGA 4lut datasets. We implement the kernels on the datasets with bit width ranging from 64-bit to 512-bit, and embedding dimension of 32. From the results, our GROOT-GPU shows a higher acceleration ratio than the other three SOFA SpMM kernels in most cases except the least complicated Booth dataset at the bit width of 64. In general, the larger the bit width of multiplier datasets and the more powerful the GPU utilized, the bigger the difference in acceleration performance. In the best case, our GROOT--GPU has the highest acceleration ratio at 10.28 for the Booth dataset, bit width 512, which is 1.667× of the second fastest MergePath--SpMM and 1.953× of the third fastest CuSPARSE. 
%at the highest, and 2.338, 2.632, 2.667 times on average
Figure~\ref{fig:eda_kernel_design} shows the comparison results
% presents the overall evaluation results, comparing the 
% of SpMM runtime acceleration ratios of 
of MergePath-SpMM, our GROOT-GPU, and CuSPARSE against GNNAdvisor, represented by the black horizontal dashed line.
% at 100\% acceleration ratio. 
The kernels are tested on the graph of the Booth Multiplier, Technology Mapping, and FPGA 4LUT datasets with bit widths ranging from 64 to 512 and an embedding dimension of 32. The kernels perform SpMM operations given the graph adjacency matrices with corresponding embeddings, and the runtime of SpMM operations are recorded by the type of kernels, the bit width of the net list which graphs describe, and the datasets where the graphs belong to.
Our GROOT-GPU demonstrates superior acceleration compared to the other three SOTA SpMM kernels in most cases.
% , except for the least complicated Booth Multiplier at a bit width of 64. 
The performance gap widens as the bit width of the multiplier datasets increases and with more powerful GPUs. GROOT-GPU achieves the highest acceleration ratio of 10.28 for the Booth dataset with a bit width of 512 on the A100 GPU, outperforming the second-fastest MergePath-SpMM by 1.67$\times$ and the third-fastest CuSPARSE by 1.95$\times$.

% \begin{figure*}[ht]
%  \centering
%    \includegraphics[width=\linewidth]{Figures/acceleration ratio v3_cropped.pdf} 

%     \caption{The runtime comparison among GROOT-GPU and SOTA GPU-based GPU Kernel designs, where the acceleration ratio of 1 from GNNAdvisor is drawn as the black dash line.}
%      \label{fig:eda_kernel_design}
% \end{figure*}
% \vspace{50pt}
\section{Conclusion}
In this paper, we introduce GROOT, an algorithm and system co-design framework that contains chip design domain knowledge, graph theory, and redesigned GPU kernels, to improve verification efficiency. We redesign nodes features utilizing
the circuit node types and the polarity of the connections between the input edges to nodes in And-Inverter Graphs (AIGs). We utilize a graph partitioning algorithm to divide the large graphs into smaller sub-graphs for fast GPU processing. After profiling EDA graph workloads, we notice their distinct distribution of high-degree and low-degree nodes and tailor the GPU kernel accordingly.
% We profile the EDA graph workloads and observe
% the uniqueness of their polarized distribution of  HD nodes and LD nodes, and redesign the GPU kernel for each.
% (HD-kernel and LD-kernel), to fit the EDA graph learning workload on a single GPU.
% combines graph partitioning and edge re-growth methods to significantly reduce memory requirements for large circuit design graphs. 
% We evaluate our framework on large circuit designs, e.g., CSA multipliers, the 7nm technology mapped CSA multipliers and Booth Multipliers. 

We compare the results with state-of-the-arts, e.g., GAMORA and ABC. 
% \cd{add kernel design.} 
Experimental results show that GROOT achieves a significant reduction in memory footprint, with high accuracy for a very large CSA multiplier, i.e., 1,024 bits with a batch size of 16. 
% Which consists of 134,103,040 nodes and 268,140,544 edges. 
We also compare GROOT with SOTA GPU-based GPU Kernel designs such as 
% will be illustrated in Fig. \ref{fig:eda_kernel_speedup}, comparing them to 
cuSPARSE, MergePath-SpMM,
% the academical frameworks such as 
and GNNAdvisor, and achieve notable runtime improvement.

\bibliographystyle{IEEEtran}
\bibliography{bib/thorat,bib/mybibliography}

\end{document}